\documentclass[runningheads]{llncs}

\usepackage{eccv}

\usepackage{eccvabbrv}

\usepackage{graphicx}
\usepackage{booktabs}
\usepackage{multicol}
\usepackage{tabularx}
\usepackage{multirow, makecell}
\usepackage{xcolor}
\usepackage{adjustbox}
\usepackage[accsupp]{axessibility}  

\usepackage{hyperref}

\usepackage{orcidlink}

\begin{document}

\title{Trajectory-aligned Space-time Tokens for Few-shot Action Recognition} 

\titlerunning{Trajectory Aligned Tokens}

\author{Pulkit Kumar\inst{1} \and Namitha Padmanabhan\inst{1} \and Luke Luo\inst{1}\and \\ Sai Saketh Rambhatla\inst{1,2}  \and Abhinav Shrivastava\inst{1}
}

\authorrunning{P.~Kumar et al.}

\institute{University of Maryland, College Park \\
\email{\{pulkit,namithap,lluo1,abhinav2\}@umd.edu} \and 
GenAI, Meta \\
\email{rssaketh@meta.com}}

 \xspaceaddexceptions{]\}}
\def\ours{PINOT\xspace}
\def\ptformer{Transformer\xspace}
\def\ptsem{TATs\xspace}
\def\cls{\texttt{CLS}\xspace}

\newcommand{\pk}[1]{{\color{ForestGreen}PK: #1}}
\newcommand{\np}[1]{{\color{VioletRed}NP: #1}}

\maketitle

\begin{abstract}
  We propose a simple yet effective approach for few-shot action recognition, emphasizing the disentanglement of motion and appearance representations. By harnessing recent progress in tracking, specifically point trajectories and self-supervised representation learning, we build trajectory-aligned tokens (TATs) that capture motion and appearance information. This approach significantly reduces the data requirements while retaining essential information. To process these representations, we use a Masked Space-time Transformer that effectively learns to aggregate information to facilitate few-shot action recognition. We demonstrate state-of-the-art results on few-shot action recognition across multiple datasets. Our project page is available \href{https://www.cs.umd.edu/~pulkit/tats}{here}.
  

\end{abstract}
\section{Introduction}
\label{sec:intro}

We present a remarkably simple yet highly effective approach for few-shot action recognition \cite{otam, zhu2018compound, mtfan, hyrsm, molo}. Unlike traditional large-scale training paradigms, the few-shot setting \cite{kliper2011one, poppe2010survey, wang2021proposal, wang2021self} demands a nuanced understanding of what constitutes an action. This requires discerning subtle cues in both motion and appearance, which are often confined to just a few pixels. We argue that, distinct from the assumptions made in large-scale action recognition, where such information can be implicitly learned in deep networks using lots of training samples, few-shot scenarios demand a more deliberate approach to model the interplay between motion and appearance cues. To address this, we propose an approach that disentangles motion and appearance representations. By leveraging recent advances in each domain, it selects a limited yet relevant subset of information to aid in learning what constitutes an action.

Our approach first models motion and then builds appearance representation aligned with this motion, which is better suited for a low-shot training paradigm with a higher signal-to-noise ratio. To model motion, we utilize the recent works on tracking a set of points across frames based on its motion\cite{doersch2023tapir, doersch2022tap, zheng2023pointodyssey, harley2022particle, xiao2024spatialtracker, cotracker}. This point-tracking paradigm for capturing motion has an inherent advantage over motion representations, such as optical flow and frame differences. First, they can persist through occlusions and capture multi-frame context, resulting in longer-term and robust motion information; and second, the set of points being tracked are often orders of magnitude smaller than a number of pixels, resulting in fewer data (\,  i.e., point trajectories) with minimal loss of information. Next, our approach is to build appearance representations aligned with these point tracks or trajectories to capture semantics. Towards this, we leverage recent advances in self-supervised methods (\eg, DINOv2~\cite{dinov2}) that are pre-trained on large datasets resulting in general-purpose vision features. The DINO family of models is particularly suited for this approach due to their self-distillation framework, which encourages consistency between different views of the same image and results in semantic representations that robustly and efficiently capture the variations introduced by actions. Concretely, we select trajectory-aligned tokens given the point trajectories and space-time DINO tokens, which effectively capture both motion and appearance cues. 

Finally, we use a masked space-time Transformer architecture\cite{arnab2021vivit,timesformer} that operates on these trajectory-aligned space-time tokens and aggregates information for each point over time and between different trajectories. The output representation of this Transformer, which captures both motion and appearance information, is then used for few-shot learning in a standard metric learning formulation~\cite{otam, trx, hyrsm, hyrsm++, hcl, bishay2019tarn, zhang2021learning, zhang2020few, wang2021semantic, ni2022multimodal}. In particular, we use the bidirectional Mean Hausdorff Metric (Bi-MHM) from~\cite{hyrsm} to analyze semantic correlations across videos and align frames to perform few-shot recognition. To summarize, our key contributions are:
\begin{itemize}
    \item We propose a simple yet effective approach for few-shot action recognition that disentangles motion and appearance representations.
    \item We leverage recent advances in tracking (\ie, point trajectories) and self-supervised representation learning (\eg, DINO) to get trajectory-aligned tokens that result in drastically less data with minimal loss of information. 
    \item We propose a Masked Space-time Transformer to learn from trajectory-aligned features and aggregate information for few-shot action recognition. 
\end{itemize}

\section{Related Work}
\label{sec:related_work}

\subsubsection{Few-shot action recognition.}
Video contains rich spatio-temporal context interactions and temporal variations \cite{kliper2011one, poppe2010survey, wang2021proposal, wang2021self}. Existing methods for few-shot action recognition predominantly apply the metric-based learning paradigm but with different focuses. Some methods focus on feature representation enhancement and spatio-temporal modelling \cite{strm, sloshnet, hyrsm, mtfan, zhu2018compound, zhu2020label}. For instance, CMN \cite{zhu2018compound} integrates a memory structure and multi-saliency algorithm to encode video representations.MTFAN \cite{mtfan} and GgHM \cite{gghm} prioritize task-specific feature learning with MTFAN using a motion encoder to integrate global motion patterns and GgHM optimizing intra- and inter-class feature correlations explicitly. SloshNet \cite{sloshnet} employs a feature fusion module to combine low- and high-level spatial features, and a long-term temporal modelling module captures global temporal relations from spatial appearance features. MoLo \cite{molo} introduces a motion autodecoder with a long-short contrastive objective to extract motion dynamics explicitly. Other methods focus on designing effective metric learning strategies \cite{otam, trx, hyrsm, hyrsm++, hcl, bishay2019tarn, zhang2021learning, zhang2020few, wang2021semantic, ni2022multimodal, fu2020depth}. OTAM \cite{otam} uses a differentiable dynamic time warping algorithm \cite{muller2007dynamic} with an ordered temporal alignment for estimating the distance between the query video and the support set videos. ARN \cite{zhang2020few} introduces a self-supervised permutation invariant strategy and learns long-range temporal dependencies by building on a 3D backbone \cite{tran2015learning}. TRX \cite{trx} matches each query sub-sequence with all sub-sequences in the support set to classify actions. HyRSM \cite{hyrsm, hyrsm++} uses hybrid relation modelling and introduces the bidirectional mean Hausdorff metric (Bi-MHM), which is adopted by several other works. ITANET \cite{zhang2021learning} and STRM \cite{strm} introduce joint spatio-temporal modelling techniques for robust few-shot matching. Nguyen et al. \cite{nguyen2022inductive} and Huang et al. \cite{huang2022compound} utilize multiple similarity functions to compare videos accurately. On the other hand, our method aims to leverage advancements in point-tracking literature and image representation learning literature to tackle few-shot action recognition.

\subsubsection{Point Tracking.}
TAP-Vid \cite{doersch2022tap} introduced the problem of tracking any physical point in a video and proposed the TAP benchmark with a simple baseline method, TAPNet. However, this method cannot track occluded points. PIPs \cite{harley2022particle} builds on the classic concept of ``particle video''~\cite{sand2008particle} that acts as a bridge between local feature matching \cite{bay2006surf, lowe2004distinctive, detone2018superpoint, shi1994good, tomasi1991detection} and optical flow~\cite{neoral2024mft, teed2020raft, dosovitskiy2015flownet, xu2017accurate, zhang2021separable, shi2023videoflow} based methods to leverage long-range temporal priors in video motion estimation. PIPs process videos in fixed-sized temporal windows and track points through occlusions, but they lose the target for longer occlusions that extend beyond the window size. TAPIR \cite{doersch2023tapir} integrates the global matching strategy from TAPNet with the refinement step of PIPs to enhance tracking accuracy. PointOdyssey~\cite{zheng2023pointodyssey} addresses long-term tracking with PIPs++, a modification of PIPs that significantly extends its temporal receptive field, and introduces a benchmark for the task of fine-grained long-term tracking. OmniMotion \cite{wang2023tracking} also addresses per-pixel per-frame trajectory estimation by ensuring global consistency of estimated motion for robust tracking through occlusions.DOT~\cite{moing2023dense} proposes a faster and more memory-efficient dense point tracking algorithm for use in practice. While the above methods estimate individual point trajectories independently, CoTracker \cite{cotracker} proposes to jointly estimate point trajectories to take advantage of the correlation between points, resulting in improved performance. We adopt the point representation of motion information in our work. Specifically, we integrate CoTracker in our method as it can efficiently and jointly track multiple points.

\subsubsection{Image Representation Learning.}

In image representation learning, two key approaches are intra-image self-supervised training and discriminative self-supervised learning. Intra-image methods utilize pretext tasks like predicting image patches \cite{doersch2015unsupervised}, re-colorizing \cite{zhang2016colorful}, or inpainting \cite{pathak2016context} to learn features. Patch-based architectures like ViTs have revived inpainting for pre-training \cite{he2022masked, bao2021beit, el2021large}, with masked auto-encoders (MAEs) demonstrating effectiveness in learning transferable features \cite{he2022masked, tong2022videomae}. On the other hand, Discriminative self-supervised learning focuses on learning features through discriminative signals between images or groups of images. This approach has evolved from early deep learning work \cite{goyal2021self} to instance classification methods \cite{alexey2016discriminative, bojanowski2017unsupervised, wu2018unsupervised} and further improvements based on instance-level objectives \cite{he2020momentum, henaff2020data, chen2021exploring, caron2021emerging} or clustering \cite{caron2018deep, asano2019self, caron2020unsupervised}.
We leverage DINOv2 \cite{dinov2} for feature extraction in our work on few-shot action recognition. DINO \cite{caron2021emerging} and DINOv2 stand out for their ability to learn rich, transferable visual representations without relying on labels. DINO introduced a self-distillation framework for learning semantically meaningful features without labels. DINOv2 builds on this, offering improved scalability and applicability across tasks. By employing DINOv2, we aim to harness the advancements in self-supervised learning for effective feature extraction in our pipeline. 
\section{Method}
\label{sec:method}
In this section we describe the details of our approach. We begin by refreshing a few preliminaries in Sec.~\ref{subsec:prelim} that we used in to build our method and then proceed to describe our method in Sec.~\ref{subsec:method}

\subsection{Preliminaries}\label{subsec:prelim}
\subsubsection{Task Details.}
In the traditional supervised learning setting,  the model is trained and tested on the same set of classes. However, in the few-shot regime, the model is trained on a specific set of base classes, denoted as $C_\text{base}$, and then evaluated on a separate set of \emph{novel} classes, $C_\text{novel}$. There is no overlap in the classes between the base and novel classes, i.e., $C_\text{base} \cap C_\text{novel} = \phi$.  For few-shot action recognition, prior works \cite{otam, trx, hyrsm, molo} typically adopt the \emph{N-way K-Shot} episode-based meta-learning strategy \cite{vinyals2016matching}. Each episode comprises a support set $S$ consisting of $N$ classes with $K$ labelled samples per class and a query set $Q$ containing query samples to be classified into one of the $N$ classes. The objective of few-shot action recognition is to learn to predict the label of a sample video from the query set using information from the support set. For training, the models train for a certain number of episodes consisting of the training data. To report the final numbers, 10000 random episodes are formed from the test data and average accuracy is reported.

\subsubsection{Point Details.}
Given an RGB video with $T$ frames and a point initialised at frame $q$ with coordinates $(x_q, y_q)$, where $q\in [1, T)$, a point tracker \cite{cotracker} tracks the point's movement across frames, yielding outputs at each time step, $(x_t, y_t)$, where $t \in [q, T]$.

\subsubsection{DINO.}
Vision Transformer (ViT) is a transformer-based model designed, where convolutions are substituted with self-attention mechanisms. The input image is divided into fixed-size patches, $\mathbf{x}_p \in \mathbb{R}^{N \times C}$, where $M$ signifies the number of patches and $C$ denotes the patch dimensions. The resulting output is $\mathbf{x}_o \in \mathbb{R}^{M \times D}$, where $D$ is the hidden dimension of the network. By leveraging ViTs, DINO~\cite{dinov2} integrates self-training and knowledge distillation without relying explicitly on labels, thereby facilitating self-supervised learning. Using a teacher-student network, DINO synthesizes global and local views from an image, with the teacher distilling its insights to inform the student network. This approach enables DINO to encapsulate a significant amount of semantic information.

\begin{figure*}[!t]
    \centering
    \includegraphics[width=0.95\linewidth]{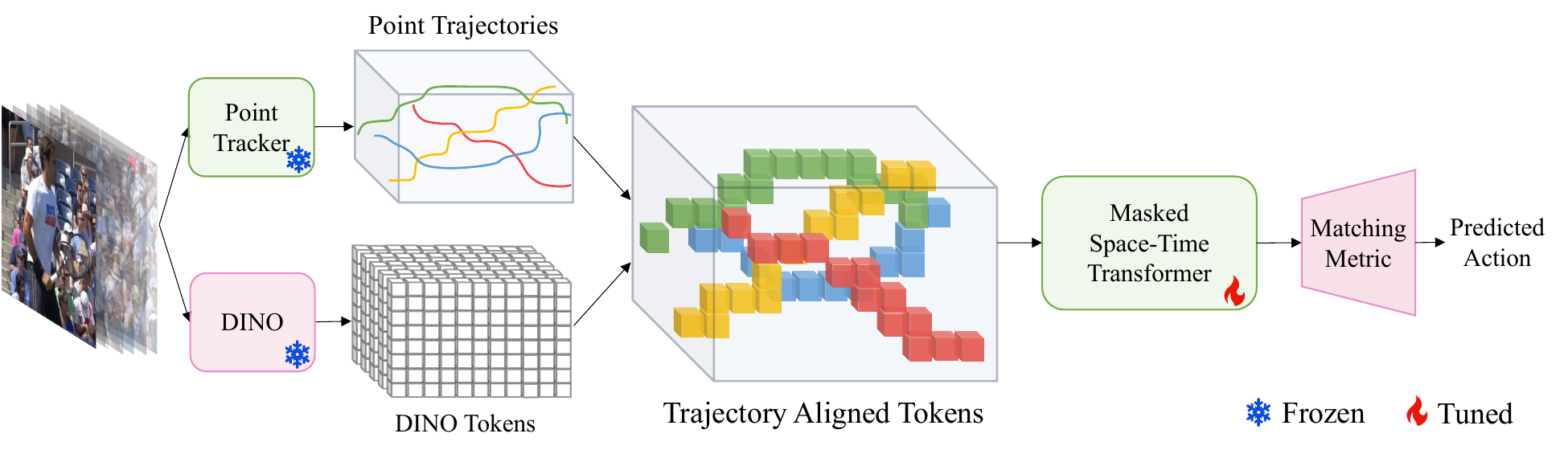}
    \caption{\textbf{Overview of our method.} We take in video frames as input and extract point trajectories and DINO patch tokens using a Point Tracker and DINO respectively. These trajectories and tokens are then aligned using a grid sampler to form trajectory-aligned tokens (TATs). Finally, we pass the TATs through a masked space-time transformer and use a matching metric on the output embedding to predict the query action.}
    \label{fig:method}
\end{figure*}

\subsection{Method}\label{subsec:method}

We first provide an overview of our method and then give more details about each component. As shown in Fig.\ref{fig:method}, given a video, we sample and track a set of points on a uniform grid using a point tracker. Simultaneously, patch tokens are extracted from each frame using DINOv2\cite{dinov2}, and the patch token corresponding to each point's location serves as its semantic descriptor. The semantic information of each point, combined with its corresponding patch token, forms the Trajectory Aligned Tokens (TATs), which are then processed by the Masked Space-Time Transformer. The weights of the point tracker and DINOv2 models are frozen, and the Space-Time Transformer is trained using a combination of cross-entropy and contrastive losses.

\subsubsection{Point extraction}
For an input video of dimensions $H{\times}W{\times}T$, we utilize Co-Tracker~\cite{cotracker} to extract a set of point trajectories across the video. Specifically, we uniformly sample a set of points on a grid in the first frame and track them using Co-Tracker. We initialize points uniformly across multiple frames to successfully track points on objects that appear at a later frame. This ensures that points on objects that appear after the first frame are tracked faithfully. 
To prevent tracking duplicates, we discard any new trajectory that is similar to the existing ones. Finally, $N$ trajectories are randomly sampled to form the set $P \in R^{T{\times}N} $ described by $\{ (x_{i}^t, y_{i}^t) , \forall t \in [q, T] , \forall i \in [1,N] , \forall q\in [1,T-1]  \}$. We ablate over the number of points and the sampling strategies in Section \ref{sec:ablations}.

\subsubsection{Trajectory-aligned Tokens (TATs)}
The spatio-temporal points in $P$, extracted from the video, lack semantic information. To enrich these points with semantics, we leverage the DINOv2 patch tokens extracted from each frame. 
We extract DINO patch tokens for all the frames in the video denoted as $\mathbf{x}_{\text{DINO}} \in \mathbb{R}^{T \times M \times D}$, where $M$ is the number of patches and $D$ is the dimension of the features.
For a point $p \in P$ at frame $q$, we use the feature of the patch token at this location as its feature descriptor.
We employ a grid sampler, that operates on $P$ and $\mathbf{x}_{\text{DINO}}$, and uses nearest neighbor interpolation to obtain $\mathbf{x}_{\text{\ptsem}} \in \mathbb{R}^{T \times N \times D}$. We term the points equipped with semantic information as  the trajectory-aligned tokens (\ptsem).

\subsubsection{Masked Space-Time Transformer.} 
Given the \ptsem, our objective is twofold: 1) to assimilate the information of each point along its trajectory, and 2) to model inter-point relationships. To achieve this, we employ a transformer equipped with the space-time attention mechanism \cite{timesformer, arnab2021vivit}. The temporal attention module helps gather point-specific information along their trajectories, while the spatial attention mechanism aggregates information across all points within each frame.
To consolidate the spatio-temporal information, an additional [\cls] token is appended to the input.

The \ptformer produces an embedding $\Tilde{f} \in \mathbb{R}^{T*N+1{\times}D}$. Previous approaches aggregate information along the spatial dimension while retaining the temporal dimension intact. Similarly, our method aggregates point information at each frame, resulting in an embedding $f \in \mathbb{R}^{T{\times}D}$. Additionally, the [\cls] token undergoes separate processing through a head layer, producing an embedding $f^{\cls} \in \mathbb{R}^{1 \times C_\text{base}}$, where $C_\text{base}$ denotes the number of base classes.  

It is important to note that not all points utilized by TATs are initialized from the initial frame; some points are initialized in subsequent frames. To facilitate processing by the transformer, we integrate a masked attention module. When considering a point initialized at time $t$, the trajectory of the point before time $t$ will be masked out during attention calculation.


\subsubsection{Set matching metric and losses.} 
The embeddings $f$ and $f^{\cls}$ are extracted for all samples in the support and the query set as described above. To classify a query sample using support samples, we adopt the set matching strategy utilized in previous methods \cite{molo,hyrsm,hyrsm++}, which employ the bidirectional Mean Hausdorff Metric (Bi-MHM) \cite{hyrsm}. Our approach also incorporates losses akin to those used in \cite{molo}. Specifically, we apply a cross-entropy loss on the [\cls] token to capture global class information and use a contrastive loss based on the Bi-MHM metric. For the equations, we refer to \cite{molo} and \cite{hyrsm}.

\section{Experiments}
This section begins by detailing the experimental configurations of our method in Section \ref{sec:exp_setup}. We then compare our approach with established state-of-the-art methods across several widely-used benchmarks in Section~\ref{sec:comp_sota}. Additionally, in Section~\ref{sec:ablations}, we provide ablations of our method, explaining the various design decisions made. Finally, Section~\ref{sec:qual_res} presents qualitative results that compare our method with previous works.

\subsection{Experimental setup}
\label{sec:exp_setup}

\subsubsection{Datasets.} To demonstrate the efficacy of our approach, we evaluate the few-shot splits of Something-Something~\cite{ssv2}, Kinetics~\cite{kinetics}, UCF101~\cite{ucf}, and HMDB51~\cite{hmdb} adopting the same splits as previous works~\cite{gghm,molo,hyrsm,hyrsm++} for a fair comparison. Two splits of Something-Something were introduced~\cite{otam} - SSv2-Small and SSv2-Full. SSv2-Full includes all classes, while SSv2-Small has 100 samples per class. For Kinetics, we used the split information from~\cite{zhu2018compound}. For UCF101 and HMDB51, we used the splits from~\cite{mtfan,zhang2020few}. Additionally, to evaluate our method on fine-grained actions, we created a few-shot split of the Fine-Gym~\cite{finegym} dataset and reported its results.

\subsubsection{Implementation details.} For extraction of points, we used Co-Tracker\cite{cotracker} for its ability to jointly and efficiently track multiple points. Since only the information of points is being used from the point tracker and they are \emph{not} fine tuned, any other methods \cite{harley2022particle,zheng2023pointodyssey,wang2023tracking,doersch2023tapir} could be used also. The points were extracted at 10 FPS. DINOv2\cite{dinov2} was used for extracting the DINO features. Following previous methods~\cite{gghm,molo,hyrsm,hyrsm++}, eight frames ($T=8$) were uniformly sampled for every video. Note that the frame-wise DINO features and the points used by our method correspond to the sampled frames. As for our transformer architecture similar to \cite{molo} was adopted. Only the transformer is trained in our setup, while the rest of the modules are frozen that leads to significant drop in training parameters as compared to previous methods; discussed in detail in the supplementary. Furthermore, we used, similar training and inference strategies were used in previous works \cite{gghm,molo,hyrsm,hyrsm++}. Final metric reported is the mean accuracy across 10,000 random episodes from the test set.

\subsection{Comparison with state-of-the-art methods}
\label{sec:comp_sota}

\begin{table}[t]
\addtolength{\tabcolsep}{2pt} 

\caption{Comparison of our method with contemporary methods of few-shot action recognition on Kinetics and SSV2 Full datasets for classification accuracy from the 1-shot up to the 5-shot settings. Entries in ``-'' mean the data is not available in published works. The best results for each are bolded. Second best results are underlined.}
\label{tab:one}
\resizebox{\textwidth}{!}{

\begin{tabular}{@{}llccccc|ccccc@{}}

\toprule

& & \multicolumn{5}{c}{\textbf{Kinetics}} & \multicolumn{5}{c}{\textbf{SSV2 Full}}  \\ \cmidrule(l){3-7} \cmidrule(l){8-12}

Method & Reference & \multicolumn{1}{l}{1-shot} & \multicolumn{1}{l}{2-shot} & \multicolumn{1}{l}{3-shot} & \multicolumn{1}{l}{4-shot} & \multicolumn{1}{l}{5-shot} & \multicolumn{1}{l}{1-shot} & \multicolumn{1}{l}{2-shot} & \multicolumn{1}{l}{3-shot} & \multicolumn{1}{l}{4-shot} & \multicolumn{1}{l}{5-shot} \\
\midrule
OTAM\cite{otam} & CVPR'20 & 72.2 & 75.9 & 78.7 & 81.9 & 84.2 & 42.8 & 49.1 & 51.5 & 52.0 & 52.3 \\
TRX\cite{trx}& CVPR'21 & 63.6 & 76.2 & 81.8 & 83.4 & 85.2 & 42.0 & 53.1 & 57.6 & 61.1 & 64.6 \\
STRM\cite{strm} & CVPR'22 & 62.9 & 76.4 & 81.1 & 83.8 & 86.7 & 43.1 & 53.3 & 59.1 & 61.7 & 68.1 \\
MTFAN\cite{mtfan} & CVPR'22 & 74.6 & - & - & - & 87.4 & 45.7 & - & - & - & 60.4 \\
HYRSM\cite{hyrsm} & CVPR'22 & 73.7 & 80.0 & 83.5 & 84.6 & 86.1 & 54.3 & 62.2 & 65.1 & 67.9 & 69.0 \\
HCL\cite{hcl} & ECCV'22 & 73.7 & 79.1 & 82.4 & 84.0 & 85.8 & 47.3 & 54.5 & 59.0 & 62.4 & 64.9 \\
Nguyen et al\cite{nguyen2022inductive} & ECCV'22 & 74.3 & - & - & - & 87.4 & 43.8 & - & - & - & 61.1 \\
Huang et al\cite{huang2022compound} & ECCV'22 & 73.3 & - & - & - & 86.4 & 49.3 & - & - & - & 66.7 \\
MoLo\cite{molo} & CVPR'23 & 74.0 & 80.4 & 83.7 & 84.7 & 85.6 & \underline{56.6} & 62.3 & \underline{67.0} & 68.5 & \underline{70.6} \\
SloshNet\cite{sloshnet} & AAAI'23 & 70.4 & - & - & - & 87.0 & 46.5 & - & - & - & 68.3 \\
GgHM\cite{gghm} & ICCV'23 & 74.9 & - & - & - & 87.4 & 54.5 & - & - & - & 69.2 \\
CCLN\cite{ccln} & PAMI'24 & \underline{75.8} & \underline{82.1} & \underline{85.0} & \underline{86.1} & \underline{87.5} & 46.0 & - & - & - & 61.3 \\
HYRSM++\cite{hyrsm++} & PR'24 & 74.0 & 80.8 & 83.9 & 85.3 & 86.4 & 55.0 & \underline{63.5} & 66.0 & \underline{68.8} & 69.8 \\
\midrule
\textbf{Ours}  & - & \textbf{81.9} & \textbf{86.5} & \textbf{89.9} & \textbf{90.6} & \textbf{91.1} & \textbf{57.7 }& \textbf{67.1 }& \textbf{70.0} & \textbf{70.6} & \textbf{74.6} \\
\bottomrule
\end{tabular}}
\end{table}

We compare our method with previous state-of-the-art methods under the 5-way $K$-shot setting. We report results with $K$ varying from 1 to 5 on Kinetics and SSV2-Full in Table \ref{tab:one}. We also report results on SSV2-small, UCF101 and HMDB51 for $K=\{1,3,5\}$ in Table \ref{table:two}. Additionally, we report results on a few-shot setting on Fine-Gym in Table \ref{table:finegym}

We observe a consistent improvement in performance across both the few-shot settings of Kinetics and SSv2 Full datasets. In the 1-shot setting of Kinetics, we observed a 6.1\% increase compared to CCLN\cite{ccln}, along with a 4.3\% enhancement in the 2-shot setting. The 3-shot setting showed an increase of 4.9\%, and the 4-shot setting demonstrated a gain of 5.3\%. Finally, in the 5-shot setting, we observed a 3.6\% improvement. In SSv2 Full, we report improvements on almost all shots compared to the previous methods. Specifically, in the 1-shot, 2-shot, and 3-shot settings, we observed gains of 2.7\%, 3.9\%, and 3\% respectively. We further noticed an improvement of 2.1\% and 4\% in 4-shot and 5-shot settings, respectively.

For SSv2 Small, we notice a 5.1\%  gain compared to \cite{hyrsm++} in 1-shot setting. For the 3-shot setting, we notice a performance gain of 6\%  as compared to MoLo\cite{molo} and for the 5-shot setting, we notice a gain of 3.8\% as compared to \cite{huang2022compound}. For UCF-101, we notice a 5.1\% gain in the 1-shot setting and a 2.4\% gain in the 3-shot setting as compared to CCLN \cite{ccln}. That said, we do see a drop of 0.8\% in the 5-shot setting. For HMDB-51, we notice a drop of 5.1\%, 4.4\% and 1.8\% as compared to the CCLN\cite{ccln}. We noticed this as a limitation to our method and have discussed this in the limitations of the paper. 

We observe similar trends in the Kinetics and SSV2 splits for the few-shot setting in Fine-Gym. We compare the performance of MoLo~\cite{molo} with our method and find that in the 1-shot setting, our method shows a performance gain of 8.5\%. In the 3-shot setting, an increase of 5.8\% is observed, and in the 5-shot setting, there is a gain of 3.1\%. This indicates the effectiveness of our method on fine-grained actions.

\subsubsection{Number of frames.}

Similar to previous approaches, we assess our performance with varying input frames and present the results in Fig \ref{fig:frames}. Focusing on the 5-way 1-shot setup in the SSV2-Full configuration, our method initially shows subpar performance with only two input frames, primarily due to its reliance on point trajectories. With just two frames, limited information leads to reduced performance. However, as the number of input frames increases, our method surpasses all previous methods, benefiting from the richer trajectory information provided by more frames.

\subsubsection{Compute analysis}
In Table \ref{tab:gflops}, we compare our model's  GFLOPS, inference time, and trainable parameters to MoLo's in a 5-way 1-shot setting. Integrating Cotrakcer and DINO increases speed and GFLOPS compared to MoLo, but the performance improvement justifies this trade-off. Additionally, our model has significantly fewer trainable parameters since neither the point tracker nor DINO are trained. Also, both can run separately offline, mitigating the impact on speed and GFLOPS.

\subsubsection{Different N-way settings.}

\begin{table}[t!]
\addtolength{\tabcolsep}{2pt} 
\caption{Comparison of our method with contemporary methods of few-shot action recognition on SSV2 Small, and UCF-101, and HMDB-51 datasets for classification accuracy on the 1-shot, 3-shot, and 5-shot settings. Entries with "-"  mean the data is not available in published works. The best results for each are bolded. Second best results are underlined.}
\label{table:two}
\resizebox{\textwidth}{!}{
\begin{tabular}{@{}llccc|ccc|ccc@{}}

\toprule
& & \multicolumn{3}{c}{\textbf{SSV2 Small}} & \multicolumn{3}{c}{\textbf{UCF-101}}  & \multicolumn{3}{c}{\textbf{HMDB-51}}  \\ \cmidrule(l){3-5} \cmidrule(l){6-8} \cmidrule(l){9-11} 
\textbf{Method} & \textbf{Reference} & \multicolumn{1}{l}{1-shot} & \multicolumn{1}{l}{3-shot} & \multicolumn{1}{l}{5-shot} & \multicolumn{1}{l}{1-shot} & \multicolumn{1}{l}{3-shot} & \multicolumn{1}{l}{5-shot} & \multicolumn{1}{l}{1-shot} & \multicolumn{1}{l}{3-shot} & \multicolumn{1}{l}{5-shot} \\
\midrule
OTAM\cite{otam}& CVPR'20& 36.4& 45.9& 48.0& 79.9& 87.0& 88.9& 54.5& 65.7& 68.0 \\
TRX~\cite{trx}& CVPR'21& 36.0& 51.9& 56.7& 78.2& 92.4& 96.1& 53.1& 66.8& 75.6 \\
STRM~\cite{strm}& CVPR'22& 37.1& 49.2& 55.3& 80.5& 92.7& \textbf{96.9}& 52.3& 67.4& 77.3 \\
MTFAN~\cite{mtfan}& CVPR'22& -& -& -& 84.8&- & 95.1&- & -& - \\
HYRSM\cite{hyrsm}& CVPR'22& 40.6&52.3& 56.1& 83.9& 93.0& 94.7& 60.3& 71.7& 76.0 \\
HCL\cite{hcl}& ECCV'22& 38.7& 49.1& 55.4& 82.5& 91.0& 93.9& 59.1& 71.2& 76.3 \\
Nguyen et al\cite{nguyen2022inductive}& ECCV'22&- & -& -& -& -&- & 59.6& -& 76.9 \\
Huang et al\cite{huang2022compound}& ECCV'22& 38.9& -& \underline{61.6}& 71.4& & 91.0& 60.1& -& 77.0 \\
MoLo\cite{molo}& CVPR'23& 42.7& \underline{52.9}& 56.4& 86.0& 93.5& 95.5& 60.8& 72.0& \underline{77.4} \\
GgHM\cite{gghm}& ICCV'23&- &- & -& 85.2&- & \underline{96.3}& 61.2&- & 76.9 \\
CCLN\cite{ccln}& PAMI'24&- &- &- & \underline{86.9}& \underline{94.2}& 96.1& \textbf{65.1}& \textbf{76.2}& \textbf{78.8} \\
HYRSM++\cite{hyrsm++}& PR'24& \underline{42.8}& 52.4& 58.0& 85.8& 93.5& 95.9& \underline{61.5}& \underline{72.7}& 76.4 \\
\midrule
\textbf{Ours} & & \textbf{47.9}& \textbf{60.0}& \textbf{64.4}& \textbf{92.0}& \textbf{96.8}& 95.5& 60.0& 71.8& 77.0 \\

\bottomrule
\end{tabular}}

\end{table}

\begin{table}[t!]

\begin{minipage}[b]{0.35\linewidth}
\addtolength{\tabcolsep}{1pt} 
\caption{Few-shot accuracy on Fine-Gym dataset.}
\label{table:finegym}
\resizebox{\linewidth}{!}{

\begin{tabular}{@{}lccc@{}}

\toprule
&  \multicolumn{3}{c}{\textbf{FineGym}}  \\
\cmidrule{2-4}
Method &  1-shot & 3-shot  & 5-shot  \\
\midrule

MoLo~\cite{molo} &  73.3 & 80.2 &84.8 \\
Ours w/o points & 77.6 &81.8 &86.4 \\
\midrule
\textbf{Ours} & \textbf{81.8} &\textbf{86.0} &\textbf{87.9} \\

\bottomrule
\end{tabular}}
\end{minipage}
\hfill
\begin{minipage}[b]{0.6\linewidth}
\addtolength{\tabcolsep}{2pt}

\caption{GFLOPS, inference time and training parameters. MST - Masked transformer}
\label{tab:gflops}

\resizebox{1\linewidth}{!}{

\begin{tabular}{@{}lcccc@{}}
\toprule

&\multicolumn{2}{c}{\textbf{GFLOPS}}
\\ \cmidrule(l){2-3} 
\textbf{Method}  & \textbf{Train} & \textbf{Test}  & \textbf{Time} & \textbf{Params}\\
\midrule
MoLo~\cite{molo}  & 252& 84 & 0.67s & 89.6M\\
Ours w/o points & 327 & 245  & 0.73s  & 10.8M\\

\midrule

Ours & 363 &  281  & 0.96s & 10.8M\\
CoTracker/Dino/MST  & (36/204/123) &  (36/204/41)  &(.23s/.24s/.49s) & (-/-/10.8M)\\

\bottomrule
\end{tabular}}

\end{minipage}

\end{table}

\begin{table}[t]
\addtolength{\tabcolsep}{2pt} 
\caption{Comparison of our method with contemporary methods of  recognition on Kinetics and SSV2 Full datasets for classification accuracy on N-way 1-shot settings. Results are reported for 5 to 10 way settings. The best results for each are bolded. Second best results are underlined.}
\label{table:three}
\resizebox{\textwidth}{!}{
\begin{tabular}{@{}lcccccc|cccccc@{}}

\toprule
& \multicolumn{6}{c}{\textbf{Kinetics}} & \multicolumn{6}{c}{\textbf{SSV2 Full}}  \\ 
\cmidrule(l){2-7} \cmidrule(l){8-13} 

\textbf{Method} &  \multicolumn{1}{l}{5-way} & \multicolumn{1}{l}{6-way} & \multicolumn{1}{l}{7-way} & \multicolumn{1}{l}{8-way} & \multicolumn{1}{l}{9-way} & \multicolumn{1}{l}{10-way}  &  \multicolumn{1}{l}{5-way} & \multicolumn{1}{l}{6-way} & \multicolumn{1}{l}{7-way} & \multicolumn{1}{l}{8-way} & \multicolumn{1}{l}{9-way} & \multicolumn{1}{l}{10-way} \\
\midrule
OTAM\cite{otam} & 72.2 & 68.7 & 66.0 & 63.0 & 61.9 & 59.0 & 42.8 & 38.6 & 35.1 & 32.3 & 30.0 & 28.2 \\
TRX\cite{trx} & 63.6 & 59.4 & 56.7 & 54.6 & 53.2 & 51.1 & 42.0 & 41.5 & 36.1 & 33.6 & 32.0 & 30.3 \\
HyRSM\cite{hyrsm} & 73.7 & 69.5 & 66.6 & 65.5 & 63.4 & 61.0 & 54.3 & 50.1 & 45.8 & 44.3 & 42.1 & 40.0 \\
MoLo\cite{molo} & \underline{74.0} & \underline{69.7} & \underline{67.4} & \underline{65.8} & \underline{63.5} & \underline{61.3} & \underline{56.6} & \underline{51.6} & \underline{48.1} & \underline{44.8} & \underline{42.5} & \underline{40.3} \\
\textbf{Ours} & \textbf{81.9} & \textbf{79.0} & \textbf{76.1} & \textbf{75.2} & \textbf{72.2} & \textbf{72.0 }& \textbf{57.7} & \textbf{55.7} &\textbf{ 52.5} & \textbf{50.0} & \textbf{47.0} & \textbf{45.8} \\
\bottomrule
\end{tabular}}

\end{table}

We also ran experiments on different \emph{N-way} settings. For both Kinetics and SSv2 Full, we run experiments on $N$ ranging from  5 to 10. We observe that on Kinetics, we get almost a 10\% increase on previous methods. The performance gain ranges from 2\%-5\% on the SSv2-Full dataset. Both of these results indicate the efficacy of our proposed method. 

 \subsection{Ablations}
 \label{sec:ablations}

In this section, we investigate the different components of our method through a series of ablation studies. We present experiments where no points are used and where points are initialized only on the first frames. Additionally, we demonstrate the impact of sampling varying numbers of points on overall performance. Moreover, we explore the effects of sampling points across different grid sizes. These ablations provide a comprehensive understanding of how each component and configuration contributes to the final performance of our method.

\subsubsection{No-point baseline and point initialisation.} To study the performance gain of our method, we created the baseline where no point information was used. Patch tokens extracted from DINO \cite{dinov2} were directly passed through the transfer and used for generating tokens. We report the number in the first row of Table \ref{table:four}. We also report performance when the points are initialised in the first frame. We report these numbers in the second row of Table \ref{table:four}. Our method adopts the uniform temporal sampling strategy where the points are reinitialised after a few frames. We report the performance in the third row of the table. 

We notice that in a 1-shot setting, the model performed worse in the first frame initialisation regime than in the baseline, where we sampled no points but a gain of performance in the uniform temporal sampling strategy. However, for both 3-shot and 5-shot kinetics settings, both point sampling strategies performed better than the baseline of no points. A similar trend in performance is noticed in SSv2-full as well as SSV2-Small. Our strategy of a) using the points and b) reinitialising the points uniformly in time performs consistently better across all settings amongst the reported datasets. This is mainly attributed to the fact that not all objects are present in the initial frame of a video; some may appear after the first frame. The uniform temporal initialisation strategy works better for catering to such objects. 

\begin{table}[t!]
\addtolength{\tabcolsep}{2pt} 
\caption{Comparison of uniformly re-initialized point sampling to baseline}
\label{table:four}
\resizebox{\textwidth}{!}{
\begin{tabular}{@{}lccc|ccc|ccc@{}}

\toprule
& \multicolumn{3}{c}{\textbf{Kinetics}} & \multicolumn{3}{c}{\textbf{SSv2-Full}}  & \multicolumn{3}{c}{\textbf{SSv2-Small}}  \\ \cmidrule(l){2-4} \cmidrule(l){5-7} \cmidrule(l){8-10} 
\textbf{Strategy} & \multicolumn{1}{l}{1-shot} & \multicolumn{1}{l}{3-shot} & \multicolumn{1}{l}{5-shot} & \multicolumn{1}{l}{1-shot} & \multicolumn{1}{l}{3-shot} & \multicolumn{1}{l}{5-shot} & \multicolumn{1}{l}{1-shot} & \multicolumn{1}{l}{3-shot} & \multicolumn{1}{l}{5-shot} \\
\midrule
No points & 80.2 & 88.8 & 90.2 & 56.2 & 68.0 & 72.5 & 44.6 & 57.7 & 62.5 \\ 
First frame only & 78.6 & 89.3 & \textbf{91.1} & 56.0 & 68.5 & 73.3 & 46.1 & 57.4 & 63.6 \\ 
Uniform temporal sampling & \textbf{81.9} & \textbf{89.9} & \textbf{91.1} & \textbf{57.7} & \textbf{70.0 }& \textbf{74.6} & \textbf{47.9} & \textbf{60.0} & \textbf{64.4} \\ 

\bottomrule
\end{tabular}}
\end{table}

\subsubsection{Number of points sampled.}
To manage memory constraints, we sample $N$ points from the set of all initialized and tracked points. It is important to note that sampling fewer points reduces the number of tokens the network processes, thereby lowering memory usage. Table \ref{table:num_points} presents our performance results with different numbers of randomly sampled points. We observe that while performance improves with an increased number of sampled points, the performance drop is not significant when the number of points is halved. For instance, sampling 128 points instead of 256 in the Kinetics dataset results in only a marginal performance decrease in both the 1-shot and 5-shot settings. Although the points are currently sampled randomly, future sampling could leverage trajectory information such as length and orientation. An effective sampling strategy could significantly improve memory efficiency as well as enhance performance. We further explore and report these results in the supplementary materials.

\begin{table}[t!]
\addtolength{\tabcolsep}{2pt} 
\center
\caption{Comparison of different number of sampled points for 1-shot, 3-shot, and 5-shot settings.}
\label{table:num_points}
\resizebox{0.9\linewidth}{!}{
\begin{tabular}{@{}lccc|ccc|ccc@{}}
\toprule
&  \multicolumn{3}{c}{\textbf{Kinetics}} & \multicolumn{3}{c}{\textbf{SSv2-Full}}  & \multicolumn{3}{c}{\textbf{SSv2-Small}}  \\ \cmidrule(l){2-4} \cmidrule(l){5-7} \cmidrule(l){8-10} 
\textbf{Points} &  \multicolumn{1}{l}{1-shot} & \multicolumn{1}{l}{3-shot} & \multicolumn{1}{l}{5-shot} & \multicolumn{1}{l}{1-shot} & \multicolumn{1}{l}{3-shot} & \multicolumn{1}{l}{5-shot} & \multicolumn{1}{l}{1-shot} & \multicolumn{1}{l}{3-shot} & \multicolumn{1}{l}{5-shot} \\
\midrule
32 &  81.0 & 86.7 & 89.4 & 53.4 & 66.2 & 71.8 & 44.5 & 57.4 & 61.5\\
64 &  81.8 & 88.0 & 89.9 & 55.5 & 68.3 & 72.5 & 45.7 & 57.7 & 62.0\\
128 &  81.6 & 89.0 & 90.5 & 56.5 & 69.1 & 73.7 & 46.8 & 59.9 & 63.9\\
256 &  \textbf{81.9} & \textbf{89.9} & \textbf{91.1} & \textbf{57.7} & \textbf{70.0} & \textbf{74.6} & \textbf{47.9} & \textbf{60.0} & \textbf{64.4}\\

\bottomrule
\end{tabular}}
\end{table}

\subsubsection{Grid size of points being sampled.}
Since no initial point information is provided, we sample them across a uniform grid of size $G$. In our previous experiments, we initialized the points uniformly on a 16-sized grid. We analyze the impact of grid size in Table \ref{table:six}, focusing on the 5-way 1-shot setting. Across the three dataset settings, we observe a slight performance drop when points are sampled on a 9-sized grid, but performance remains relatively constant when the grid size increases to 25. This finding aligns with our previous observations on ablating for the number of points. It's worth noting that as the grid size increases, so does the number of tokens and, consequently, the model's memory consumption. Hence, there exists a slight trade-off between grid size and memory consumption.

\begin{table}[t!]

\begin{minipage}[b]{0.48\linewidth}
\addtolength{\tabcolsep}{2pt} 
\caption{Classification accuracy of 5-way 1-shot setting for several grid sizes.}
\label{table:six}
\resizebox{\linewidth}{!}{

\begin{tabular}{@{}cccc@{}}

\toprule
\textbf{Grid Size} &  \textbf{Kinetics} & \textbf{SSv2-Full}  & \textbf{SSv2-Small}  \\
\midrule

\textbf{$9 \times 9$} & 81.3 &56.0 &45.7 \\
\textbf{$16 \times 16$} & 81.9 &57.7 &\textbf{47.9} \\
\textbf{$25 \times 25$} & \textbf{82.1} &\textbf{58.1} &47.7 \\

\bottomrule
\end{tabular}}
\end{minipage}
\hfill
\begin{minipage}[t]{0.48\linewidth}
\centering
    \includegraphics[width=\linewidth]{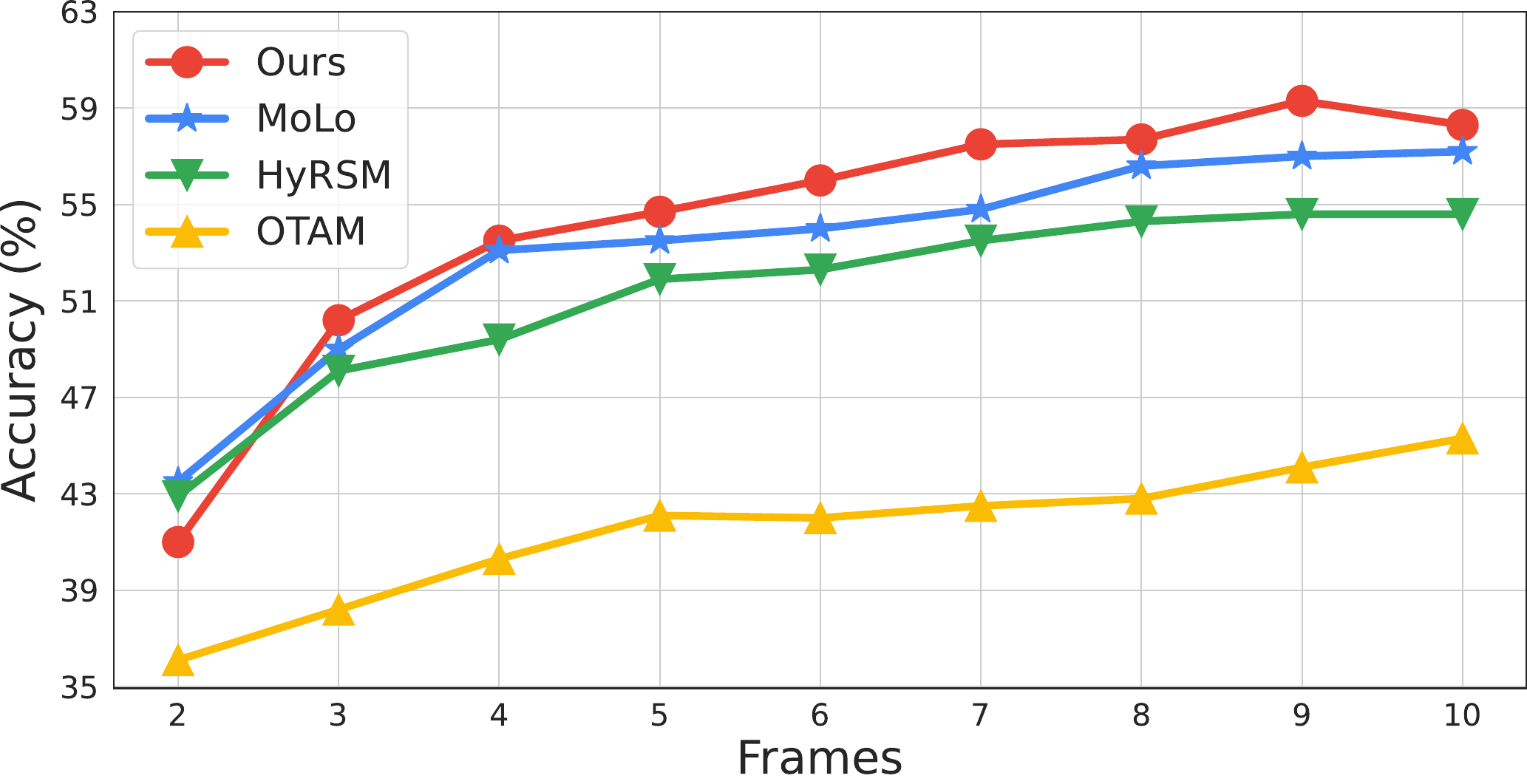}
    \captionof{figure}{Effect of  number of input frames under the 5-way 1-shot setting on SSv2-Full.}
    \label{fig:frames}

\end{minipage}

\end{table}



\subsection{Qualitative results}
\label{sec:qual_res}
\begin{figure}[!t]
\centering
\begin{subfigure}{1\textwidth}
  \centering
  \includegraphics[width=\linewidth]{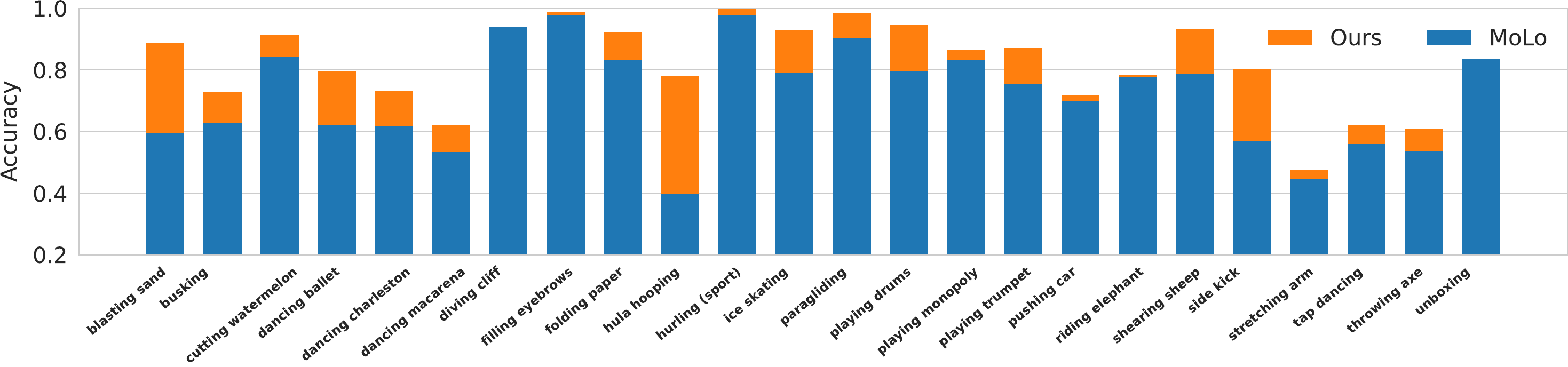}
  \subcaption{Kinetics}
  \label{fig:sub1}
\end{subfigure}%
 \hspace*{\fill}

\begin{subfigure}{1\textwidth}
  \centering
  \includegraphics[width=\linewidth]{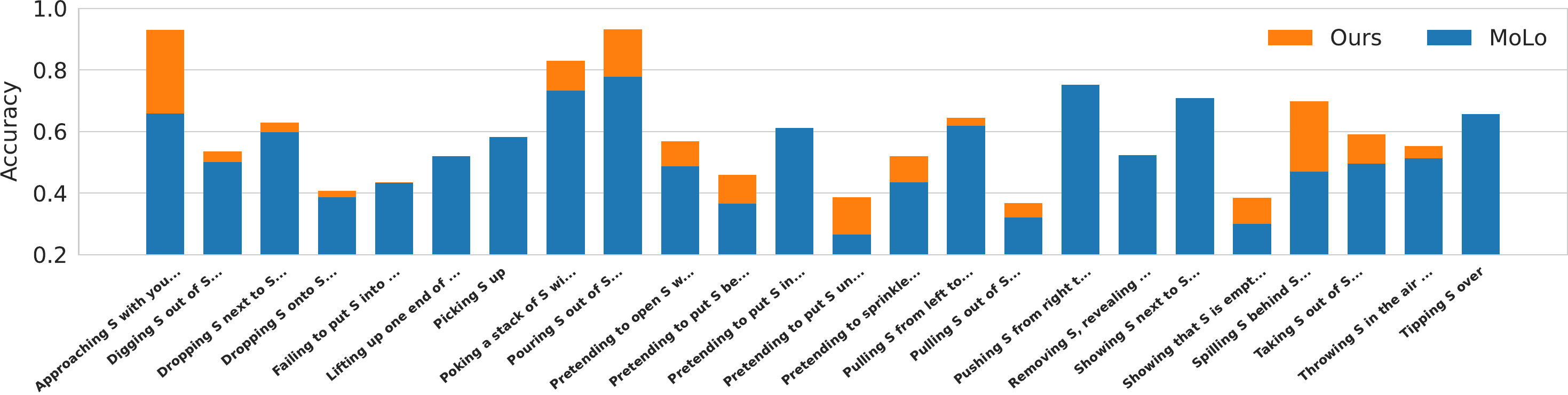}
  \subcaption{SSv2 Full}
  \label{fig:sub2}
\end{subfigure}
\begin{subfigure}{1\textwidth}
  \centering
  \includegraphics[width=\linewidth]{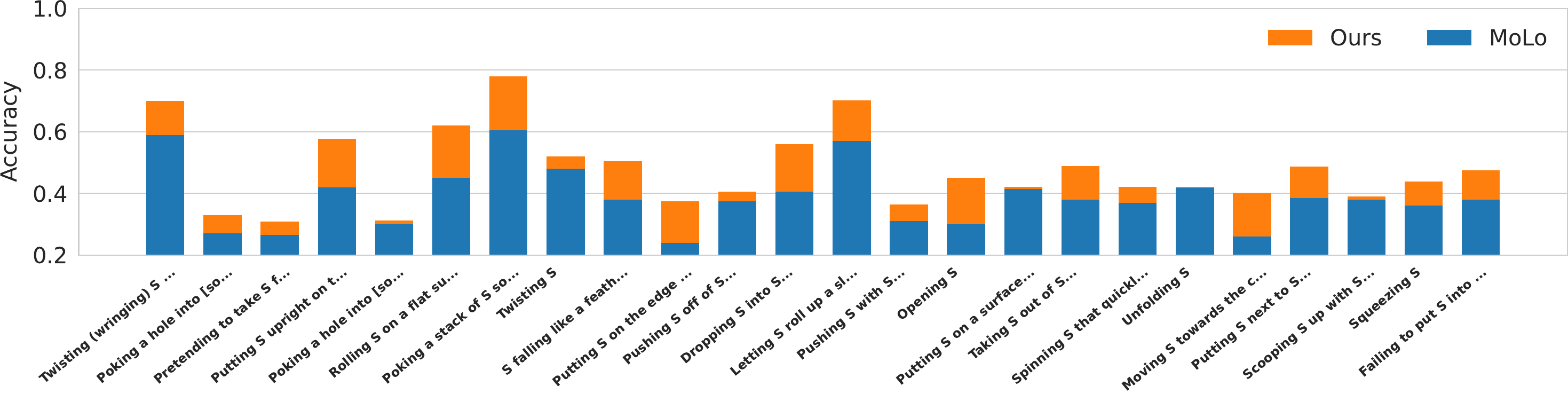}
  \subcaption{SSv2 Small}
  \label{fig:sub3}
\end{subfigure}
\caption{Quantitative analysis of 5-way 1-shot setting compared to MoLo. Top: Kinetics dataset; Middle: SSv2 Full dataset; Bottom: SSv2 Small dataset. "S" is a shorthand for "Something".}
\label{fig:acc_plot}
\end{figure}
\begin{figure*}[t!]
    \centering
    \includegraphics[width=\linewidth]{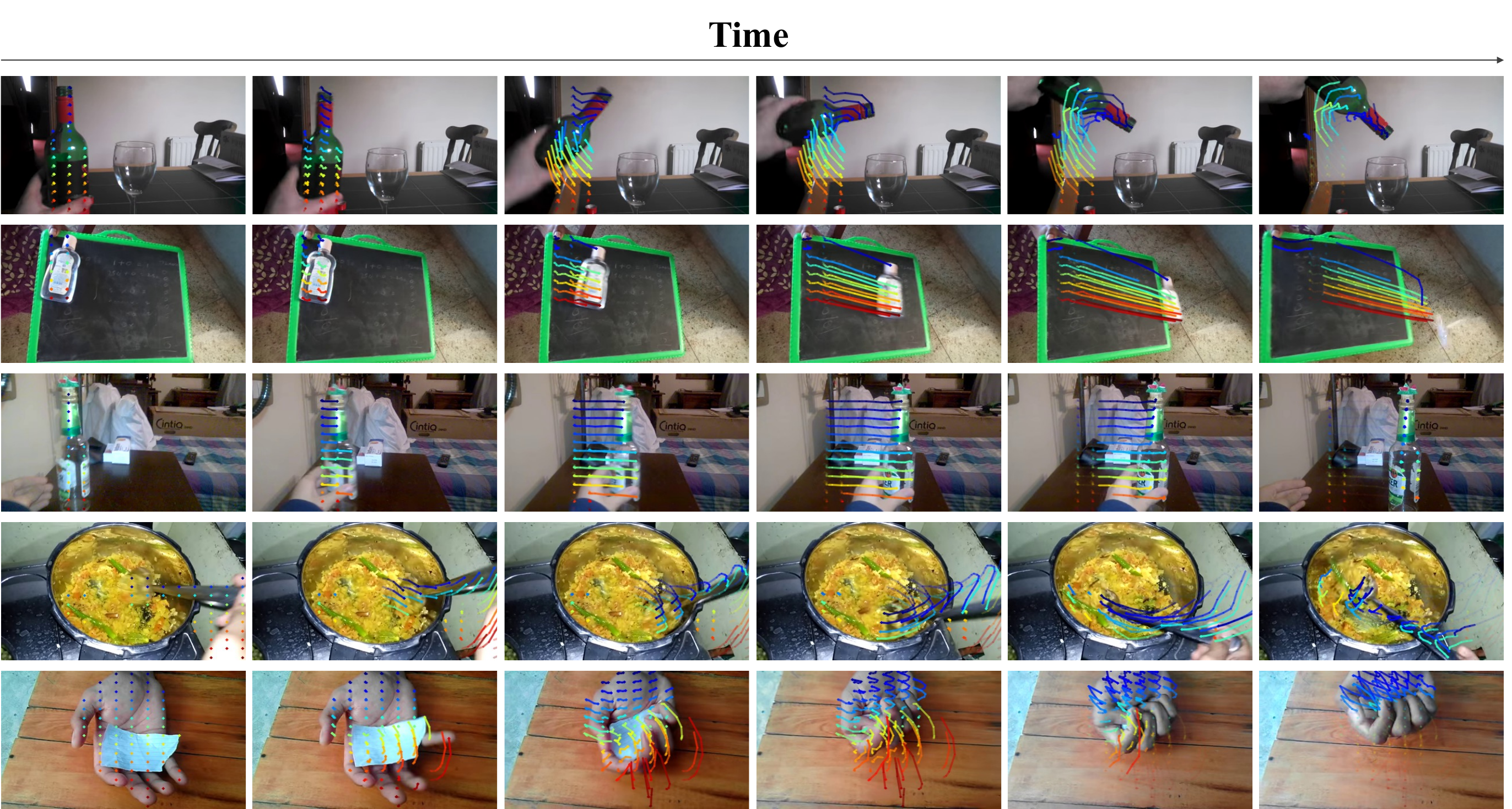}
        \caption{\textbf{Qualitative Results.} We showcase some results where our method performs better than baselines, attributed to the motion information that is clearly discernible in these samples. The examples are drawn from the Something-Something \cite{ssv2} dataset. The six columns depict the frames of each video sampled across time, and the lines denote the trajectories of tracked points. For visualization purposes, only the points on the most salient object are visualised while the background points are omitted. }
    \label{fig:qual_with_points}
\end{figure*}

Following~\cite{molo}, we analyze the class-wise accuracy performance of a 5-way 1-shot setting across the three datasets and report them in Fig \ref{fig:acc_plot}. We compare our method with MoLo~\cite{molo} across the three dataset settings. In Kinetics, we gain significant performance in high-motion classes such as hula hooping, sidekick, blasting sand, etc. But no gain was seen in classes such as cliff diving, unboxing, etc.  This limitation is disucssed in Section~\ref{sec:limitation}.

In SSv2-full, performance gains were seen in classes like ``approaching something with something'', ``pouring something out of something'', ``spilling something behind something'' showed performance improvement. However, actions like ``failing to put something into something'', ``removing something to reveal something'', ``showing something next to something'', and ``tipping something over'' did not show improvements. The limitations of our method show that the actions that track points with a persistent spatial position are more challenging to classify.

In SSv2-small, our method shows an improvement in performance for queries such as ``Putting something upright'', ``Rolling something'', ``Poking a stack of something so it collapses'', and ``Dropping something''. These classes typically feature relevant objects moving into and out of frame, a type of obfuscation that conventional methods of few-shot action recognition struggle to identify. However, classes like ``Scooping something'', ``Putting something on a surface'', and ``Unfolding something'', in which the relative motion of relevant points is noticeably lesser, did not show improvements and is the limitation of our method. 

In Fig~\ref{fig:qual_with_points}, we visualize several videos from the SSV2 dataset, highlighting the point tracking information. For clarity, we display only the points associated with the salient object, omitting the background points. However, all points, including those in the background, are utilized in our approach.

\section{Limitations}
\label{sec:limitation}
Since our methodology draws from two distinct domains, it potentially inherits their respective limitations and errors. Point tracking, for instance, proves highly susceptible to significant camera motion. In scenarios where motion is pronounced, the efficacy of point tracking diminishes, resulting in suboptimal performance for our method. Another challenge we encountered pertains to videos characterized by abrupt scene cuts, a phenomenon not prevalent in SSv2 but commonly found in Kinetics. Managing point tracking under such circumstances proves intricate. Furthermore, we observed a distinct limitation in cases where motion information provides limited assistance in action recognition (e.g., “smoking,” “talking”), as our trajectory-aligned features fail to yield discernible improvements. Improvements in point tracking methods can help solve many of the limitations faced. 

\section{Discussion and future work}

This work integrates points and their trajectories using DINOv2 patch tokens to form Trajectory-aligned Tokens (TATs). While we focused on utilizing TATs for few-shot action recognition, we believe they hold potential for other action recognition tasks as well. There are several aspects that we did not address in our current work but could be crucial for solving different tasks. Notably, our work does not account for the visibility of points, which is information provided by point trackers. Point visibility can be essential for fine-grained recognition. Additionally, we recognize the need for improved sampling strategies to select the points used by TATs. Since similar objects carry similar information, not all points on the objects are necessary for the downstream task. A more intelligent sampling strategy would enhance the efficiency of our method. We hope that future research will leverage TATs and explore innovative approaches to address other video-related tasks.

\section{Conclusion}
In conclusion, we present a straightforward yet impactful method for few-shot action recognition, focusing on separating motion and appearance representations. Leveraging advancements in tracking, particularly point trajectories and self-supervised learning, we introduce trajectory-aligned tokens (TATs) that encode both motion and appearance cues. This technique not only minimizes data needs but also preserves crucial information. Employing a Masked Space-time Transformer to process these representations, we effectively learn to consolidate information for improved few-shot action recognition. Our experimental results showcase leading performance on various datasets, underlining the efficacy of our approach. We believe that our work marks a significant step forward in the field of action recognition, offering a powerful tool for future research and practical applications.

\noindent\textbf{Acknowledgements.} We would like to thank Nirat Saini and Matthew Gwilliam for their helpful feedback while we prepared the manuscript. This work was partially supported by NSF CAREER Award (\#2238769) to AS. The authors acknowledge UMD’s supercomputing resources made available for conducting this research. The U.S. Government is authorized to reproduce and distribute reprints for Governmental purposes notwithstanding any copyright annotation thereon. The views and conclusions contained herein are those of the authors and should not be interpreted as necessarily representing the official policies or endorsements, either expressed or implied, of NSF or the U.S. Government.

\bibliographystyle{splncs04}
\bibliography{main}

\begin{thebibliography}{10}
\providecommand{\url}[1]{\texttt{#1}}
\providecommand{\urlprefix}{URL }
\providecommand{\doi}[1]{https://doi.org/#1}

\bibitem{alexey2016discriminative}
Alexey, D., Fischer, P., Tobias, J., Springenberg, M.R., Brox, T.: Discriminative unsupervised feature learning with exemplar convolutional neural networks. IEEE TPAMI  \textbf{38}(9),  1734--1747 (2016)

\bibitem{arnab2021vivit}
Arnab, A., Dehghani, M., Heigold, G., Sun, C., Lu{\v{c}}i{\'c}, M., Schmid, C.: Vivit: A video vision transformer. In: Proceedings of the IEEE/CVF international conference on computer vision. pp. 6836--6846 (2021)

\bibitem{asano2019self}
Asano, Y.M., Rupprecht, C., Vedaldi, A.: Self-labelling via simultaneous clustering and representation learning. arXiv preprint arXiv:1911.05371  (2019)

\bibitem{bao2021beit}
Bao, H., Dong, L., Piao, S., Wei, F.: Beit: Bert pre-training of image transformers. arXiv preprint arXiv:2106.08254  (2021)

\bibitem{bay2006surf}
Bay, H., Tuytelaars, T., Van~Gool, L.: Surf: Speeded up robust features. In: Computer Vision--ECCV 2006: 9th European Conference on Computer Vision, Graz, Austria, May 7-13, 2006. Proceedings, Part I 9. pp. 404--417. Springer (2006)

\bibitem{timesformer}
Bertasius, G., Wang, H., Torresani, L.: Is space-time attention all you need for video understanding? In: Proceedings of the International Conference on Machine Learning (ICML) (July 2021)

\bibitem{bishay2019tarn}
Bishay, M., Zoumpourlis, G., Patras, I.: Tarn: Temporal attentive relation network for few-shot and zero-shot action recognition. arXiv preprint arXiv:1907.09021  (2019)

\bibitem{bojanowski2017unsupervised}
Bojanowski, P., Joulin, A.: Unsupervised learning by predicting noise. In: International Conference on Machine Learning. pp. 517--526. PMLR (2017)

\bibitem{otam}
Cao, K., Ji, J., Cao, Z., Chang, C.Y., Niebles, J.C.: Few-shot video classification via temporal alignment. In: Proceedings of the IEEE/CVF Conference on Computer Vision and Pattern Recognition. pp. 10618--10627 (2020)

\bibitem{caron2018deep}
Caron, M., Bojanowski, P., Joulin, A., Douze, M.: Deep clustering for unsupervised learning of visual features. In: Proceedings of the European conference on computer vision (ECCV). pp. 132--149 (2018)

\bibitem{caron2020unsupervised}
Caron, M., Misra, I., Mairal, J., Goyal, P., Bojanowski, P., Joulin, A.: Unsupervised learning of visual features by contrasting cluster assignments. Advances in neural information processing systems  \textbf{33},  9912--9924 (2020)

\bibitem{caron2021emerging}
Caron, M., Touvron, H., Misra, I., J\'egou, H., Mairal, J., Bojanowski, P., Joulin, A.: Emerging properties in self-supervised vision transformers. In: Proceedings of the International Conference on Computer Vision (ICCV) (2021)

\bibitem{kinetics}
Carreira, J., Zisserman, A.: Quo vadis, action recognition? a new model and the kinetics dataset. 2017 IEEE Conference on Computer Vision and Pattern Recognition (CVPR) pp. 4724--4733 (2017)

\bibitem{chen2021exploring}
Chen, X., He, K.: Exploring simple siamese representation learning. In: Proceedings of the IEEE/CVF conference on computer vision and pattern recognition. pp. 15750--15758 (2021)

\bibitem{detone2018superpoint}
DeTone, D., Malisiewicz, T., Rabinovich, A.: Superpoint: Self-supervised interest point detection and description. In: Proceedings of the IEEE conference on computer vision and pattern recognition workshops. pp. 224--236 (2018)

\bibitem{doersch2015unsupervised}
Doersch, C., Gupta, A., Efros, A.A.: Unsupervised visual representation learning by context prediction. In: Proceedings of the IEEE international conference on computer vision. pp. 1422--1430 (2015)

\bibitem{doersch2022tap}
Doersch, C., Gupta, A., Markeeva, L., Recasens, A., Smaira, L., Aytar, Y., Carreira, J., Zisserman, A., Yang, Y.: Tap-vid: A benchmark for tracking any point in a video. Advances in Neural Information Processing Systems  \textbf{35},  13610--13626 (2022)

\bibitem{doersch2023tapir}
Doersch, C., Yang, Y., Vecerik, M., Gokay, D., Gupta, A., Aytar, Y., Carreira, J., Zisserman, A.: Tapir: Tracking any point with per-frame initialization and temporal refinement. arXiv preprint arXiv:2306.08637  (2023)

\bibitem{dosovitskiy2015flownet}
Dosovitskiy, A., Fischer, P., Ilg, E., Hausser, P., Hazirbas, C., Golkov, V., Van Der~Smagt, P., Cremers, D., Brox, T.: Flownet: Learning optical flow with convolutional networks. In: Proceedings of the IEEE international conference on computer vision. pp. 2758--2766 (2015)

\bibitem{el2021large}
El-Nouby, A., Izacard, G., Touvron, H., Laptev, I., Jegou, H., Grave, E.: Are large-scale datasets necessary for self-supervised pre-training? arXiv preprint arXiv:2112.10740  (2021)

\bibitem{fu2020depth}
Fu, Y., Zhang, L., Wang, J., Fu, Y., Jiang, Y.G.: Depth guided adaptive meta-fusion network for few-shot video recognition. In: Proceedings of the 28th ACM International Conference on Multimedia. pp. 1142--1151 (2020)

\bibitem{goyal2021self}
Goyal, P., Caron, M., Lefaudeux, B., Xu, M., Wang, P., Pai, V., Singh, M., Liptchinsky, V., Misra, I., Joulin, A., et~al.: Self-supervised pretraining of visual features in the wild. arXiv preprint arXiv:2103.01988  (2021)

\bibitem{ssv2}
Goyal, R., Kahou, S.E., Michalski, V., Materzynska, J., Westphal, S., Kim, H., Haenel, V., Fr{\"u}nd, I., Yianilos, P.N., Mueller-Freitag, M., Hoppe, F., Thurau, C., Bax, I., Memisevic, R.: The “something something” video database for learning and evaluating visual common sense. 2017 IEEE International Conference on Computer Vision (ICCV) pp. 5843--5851 (2017)

\bibitem{harley2022particle}
Harley, A.W., Fang, Z., Fragkiadaki, K.: Particle video revisited: Tracking through occlusions using point trajectories. In: European Conference on Computer Vision. pp. 59--75. Springer (2022)

\bibitem{he2022masked}
He, K., Chen, X., Xie, S., Li, Y., Doll{\'a}r, P., Girshick, R.: Masked autoencoders are scalable vision learners. In: Proceedings of the IEEE/CVF conference on computer vision and pattern recognition. pp. 16000--16009 (2022)

\bibitem{he2020momentum}
He, K., Fan, H., Wu, Y., Xie, S., Girshick, R.: Momentum contrast for unsupervised visual representation learning. In: Proceedings of the IEEE/CVF conference on computer vision and pattern recognition. pp. 9729--9738 (2020)

\bibitem{henaff2020data}
Henaff, O.: Data-efficient image recognition with contrastive predictive coding. In: International conference on machine learning. pp. 4182--4192. PMLR (2020)

\bibitem{huang2022compound}
Huang, Y., Yang, L., Sato, Y.: Compound prototype matching for few-shot action recognition. In: European Conference on Computer Vision. pp. 351--368. Springer (2022)

\bibitem{cotracker}
Karaev, N., Rocco, I., Graham, B., Neverova, N., Vedaldi, A., Rupprecht, C.: Cotracker: It is better to track together. arXiv preprint arXiv:2307.07635  (2023)

\bibitem{kliper2011one}
Kliper-Gross, O., Hassner, T., Wolf, L.: One shot similarity metric learning for action recognition. In: Similarity-Based Pattern Recognition: First International Workshop, SIMBAD 2011, Venice, Italy, September 28-30, 2011. Proceedings 1. pp. 31--45. Springer (2011)

\bibitem{hmdb}
Kuehne, H., Jhuang, H., Garrote, E., Poggio, T.A., Serre, T.: Hmdb: A large video database for human motion recognition. 2011 International Conference on Computer Vision pp. 2556--2563 (2011)

\bibitem{lowe2004distinctive}
Lowe, D.G.: Distinctive image features from scale-invariant keypoints. International journal of computer vision  \textbf{60},  91--110 (2004)

\bibitem{moing2023dense}
Moing, G.L., Ponce, J., Schmid, C.: Dense optical tracking: Connecting the dots. arXiv preprint arXiv:2312.00786  (2023)

\bibitem{muller2007dynamic}
Muller, M.: Dynamic time warping in information retrieval for music and motion. Dynamic time warping Information retrieval for music and motion pp. 69--84 (2007)

\bibitem{neoral2024mft}
Neoral, M., {\v{S}}er{\`y}ch, J., Matas, J.: Mft: Long-term tracking of every pixel. In: Proceedings of the IEEE/CVF Winter Conference on Applications of Computer Vision. pp. 6837--6847 (2024)

\bibitem{nguyen2022inductive}
Nguyen, K.D., Tran, Q.H., Nguyen, K., Hua, B.S., Nguyen, R.: Inductive and transductive few-shot video classification via appearance and temporal alignments. In: European Conference on Computer Vision. pp. 471--487. Springer (2022)

\bibitem{ni2022multimodal}
Ni, X., Liu, Y., Wen, H., Ji, Y., Xiao, J., Yang, Y.: Multimodal prototype-enhanced network for few-shot action recognition. arXiv preprint arXiv:2212.04873  (2022)

\bibitem{dinov2}
Oquab, M., Darcet, T., Moutakanni, T., Vo, H.Q., Szafraniec, M., Khalidov, V., Fernandez, P., Haziza, D., Massa, F., El-Nouby, A., Assran, M., Ballas, N., Galuba, W., Howes, R., Huang, P.Y.B., Li, S.W., Misra, I., Rabbat, M.G., Sharma, V., Synnaeve, G., Xu, H., J{\'e}gou, H., Mairal, J., Labatut, P., Joulin, A., Bojanowski, P.: Dinov2: Learning robust visual features without supervision. ArXiv  \textbf{abs/2304.07193} (2023)

\bibitem{pathak2016context}
Pathak, D., Krahenbuhl, P., Donahue, J., Darrell, T., Efros, A.A.: Context encoders: Feature learning by inpainting. In: Proceedings of the IEEE conference on computer vision and pattern recognition. pp. 2536--2544 (2016)

\bibitem{trx}
Perrett, T., Masullo, A., Burghardt, T., Mirmehdi, M., Damen, D.: Temporal-relational crosstransformers for few-shot action recognition. In: Proceedings of the IEEE/CVF conference on computer vision and pattern recognition. pp. 475--484 (2021)

\bibitem{poppe2010survey}
Poppe, R.: A survey on vision-based human action recognition. Image and vision computing  \textbf{28}(6),  976--990 (2010)

\bibitem{sand2008particle}
Sand, P., Teller, S.: Particle video: Long-range motion estimation using point trajectories. International journal of computer vision  \textbf{80},  72--91 (2008)

\bibitem{finegym}
Shao, D., Zhao, Y., Dai, B., Lin, D.: Finegym: A hierarchical video dataset for fine-grained action understanding. In: IEEE Conference on Computer Vision and Pattern Recognition (CVPR) (2020)

\bibitem{shi1994good}
Shi, J., et~al.: Good features to track. In: 1994 Proceedings of IEEE conference on computer vision and pattern recognition. pp. 593--600. IEEE (1994)

\bibitem{shi2023videoflow}
Shi, X., Huang, Z., Bian, W., Li, D., Zhang, M., Cheung, K.C., See, S., Qin, H., Dai, J., Li, H.: Videoflow: Exploiting temporal cues for multi-frame optical flow estimation. arXiv preprint arXiv:2303.08340  (2023)

\bibitem{ucf}
Soomro, K., Zamir, A., Shah, M.: Ucf101: A dataset of 101 human actions classes from videos in the wild. ArXiv  \textbf{abs/1212.0402} (2012)

\bibitem{teed2020raft}
Teed, Z., Deng, J.: Raft: Recurrent all-pairs field transforms for optical flow. In: Computer Vision--ECCV 2020: 16th European Conference, Glasgow, UK, August 23--28, 2020, Proceedings, Part II 16. pp. 402--419. Springer (2020)

\bibitem{strm}
Thatipelli, A., Narayan, S., Khan, S., Anwer, R.M., Khan, F.S., Ghanem, B.: Spatio-temporal relation modeling for few-shot action recognition. In: Proceedings of the IEEE/CVF Conference on Computer Vision and Pattern Recognition. pp. 19958--19967 (2022)

\bibitem{tomasi1991detection}
Tomasi, C., Kanade, T.: Detection and tracking of point. Int J Comput Vis  \textbf{9}(137-154), ~3 (1991)

\bibitem{tong2022videomae}
Tong, Z., Song, Y., Wang, J., Wang, L.: Videomae: Masked autoencoders are data-efficient learners for self-supervised video pre-training. Advances in neural information processing systems  \textbf{35},  10078--10093 (2022)

\bibitem{tran2015learning}
Tran, D., Bourdev, L., Fergus, R., Torresani, L., Paluri, M.: Learning spatiotemporal features with 3d convolutional networks. In: Proceedings of the IEEE international conference on computer vision. pp. 4489--4497 (2015)

\bibitem{vinyals2016matching}
Vinyals, O., Blundell, C., Lillicrap, T., Wierstra, D., et~al.: Matching networks for one shot learning. Advances in neural information processing systems  \textbf{29} (2016)

\bibitem{wang2023tracking}
Wang, Q., Chang, Y.Y., Cai, R., Li, Z., Hariharan, B., Holynski, A., Snavely, N.: Tracking everything everywhere all at once. arXiv preprint arXiv:2306.05422  (2023)

\bibitem{wang2021proposal}
Wang, X., Qing, Z., Huang, Z., Feng, Y., Zhang, S., Jiang, J., Tang, M., Gao, C., Sang, N.: Proposal relation network for temporal action detection. arXiv preprint arXiv:2106.11812  (2021)

\bibitem{molo}
Wang, X., Zhang, S., Qing, Z., Gao, C., Zhang, Y., Zhao, D., Sang, N.: Molo: Motion-augmented long-short contrastive learning for few-shot action recognition. In: Proceedings of the IEEE/CVF Conference on Computer Vision and Pattern Recognition. pp. 18011--18021 (2023)

\bibitem{wang2021self}
Wang, X., Zhang, S., Qing, Z., Shao, Y., Gao, C., Sang, N.: Self-supervised learning for semi-supervised temporal action proposal. In: Proceedings of the IEEE/CVF Conference on Computer Vision and Pattern Recognition. pp. 1905--1914 (2021)

\bibitem{hyrsm}
Wang, X., Zhang, S., Qing, Z., Tang, M., Zuo, Z., Gao, C., Jin, R., Sang, N.: Hybrid relation guided set matching for few-shot action recognition. In: Proceedings of the IEEE/CVF Conference on Computer Vision and Pattern Recognition. pp. 19948--19957 (2022)

\bibitem{hyrsm++}
Wang, X., Zhang, S., Qing, Z., Zuo, Z., Gao, C., Jin, R., Sang, N.: Hyrsm++: Hybrid relation guided temporal set matching for few-shot action recognition. Pattern Recognition  \textbf{147},  110110 (2024)

\bibitem{ccln}
Wang, X., Yan, Y., Hu, H.M., Li, B., Wang, H.: Cross-modal contrastive learning network for few-shot action recognition. IEEE Transactions on Image Processing  (2024)

\bibitem{wang2021semantic}
Wang, X., Ye, W., Qi, Z., Zhao, X., Wang, G., Shan, Y., Wang, H.: Semantic-guided relation propagation network for few-shot action recognition. In: Proceedings of the 29th ACM International Conference on Multimedia. pp. 816--825 (2021)

\bibitem{mtfan}
Wu, J., Zhang, T., Zhang, Z., Wu, F., Zhang, Y.: Motion-modulated temporal fragment alignment network for few-shot action recognition. In: Proceedings of the IEEE/CVF Conference on Computer Vision and Pattern Recognition. pp. 9151--9160 (2022)

\bibitem{wu2018unsupervised}
Wu, Z., Xiong, Y., Yu, S.X., Lin, D.: Unsupervised feature learning via non-parametric instance discrimination. In: Proceedings of the IEEE conference on computer vision and pattern recognition. pp. 3733--3742 (2018)

\bibitem{xiao2024spatialtracker}
Xiao, Y., Wang, Q., Zhang, S., Xue, N., Peng, S., Shen, Y., Zhou, X.: Spatialtracker: Tracking any 2d pixels in 3d space. In: Proceedings of the IEEE/CVF Conference on Computer Vision and Pattern Recognition. pp. 20406--20417 (2024)

\bibitem{sloshnet}
Xing, J., Wang, M., Mu, B., Liu, Y.: Revisiting the spatial and temporal modeling for few-shot action recognition. In: AAAI Conference on Artificial Intelligence (2023), \url{https://api.semanticscholar.org/CorpusID:255999953}

\bibitem{gghm}
Xing, J., Wang, M., Ruan, Y., Chen, B., Guo, Y., Mu, B., Dai, G., Wang, J., Liu, Y.: Boosting few-shot action recognition with graph-guided hybrid matching. In: Proceedings of the IEEE/CVF International Conference on Computer Vision. pp. 1740--1750 (2023)

\bibitem{xu2017accurate}
Xu, J., Ranftl, R., Koltun, V.: Accurate optical flow via direct cost volume processing. In: Proceedings of the IEEE Conference on Computer Vision and Pattern Recognition. pp. 1289--1297 (2017)

\bibitem{zhang2021separable}
Zhang, F., Woodford, O.J., Prisacariu, V.A., Torr, P.H.: Separable flow: Learning motion cost volumes for optical flow estimation. In: Proceedings of the IEEE/CVF international conference on computer vision. pp. 10807--10817 (2021)

\bibitem{zhang2020few}
Zhang, H., Zhang, L., Qi, X., Li, H., Torr, P.H., Koniusz, P.: Few-shot action recognition with permutation-invariant attention. In: Computer Vision--ECCV 2020: 16th European Conference, Glasgow, UK, August 23--28, 2020, Proceedings, Part V 16. pp. 525--542. Springer (2020)

\bibitem{zhang2016colorful}
Zhang, R., Isola, P., Efros, A.A.: Colorful image colorization. In: Computer Vision--ECCV 2016: 14th European Conference, Amsterdam, The Netherlands, October 11-14, 2016, Proceedings, Part III 14. pp. 649--666. Springer (2016)

\bibitem{zhang2021learning}
Zhang, S., Zhou, J., He, X.: Learning implicit temporal alignment for few-shot video classification. arXiv preprint arXiv:2105.04823  (2021)

\bibitem{hcl}
Zheng, S., Chen, S., Jin, Q.: Few-shot action recognition with hierarchical matching and contrastive learning. In: European Conference on Computer Vision. pp. 297--313. Springer (2022)

\bibitem{zheng2023pointodyssey}
Zheng, Y., Harley, A.W., Shen, B., Wetzstein, G., Guibas, L.J.: Pointodyssey: A large-scale synthetic dataset for long-term point tracking. In: Proceedings of the IEEE/CVF International Conference on Computer Vision. pp. 19855--19865 (2023)

\bibitem{zhu2018compound}
Zhu, L., Yang, Y.: Compound memory networks for few-shot video classification. In: Proceedings of the European Conference on Computer Vision (ECCV). pp. 751--766 (2018)

\bibitem{zhu2020label}
Zhu, L., Yang, Y.: Label independent memory for semi-supervised few-shot video classification. IEEE Transactions on Pattern Analysis and Machine Intelligence  \textbf{44}(1),  273--285 (2020)

\end{thebibliography}
\clearpage
\maketitlesupplementary

\section{Comparison with previous methods (extended)}

In this section, we compare the trainable parameters utilized in our method with those of previous techniques. The parameter counts are detailed in Table \ref{table:params}, highlighting a substantial reduction in our approach compared to earlier methods. Specifically, while all other methods reported fine-tuning their backbone networks, our method requires training only for the space-time transformer, omitting the need for any backbone fine-tuning. This streamlined approach enhances efficiency in training without compromising in performance.

\section{Ablations (extended)}

In this section, we delve into various design decisions for our method and present the results of each. Firstly, we showcase the outcomes achieved by employing different point trackers alongside our method. Following this, we elaborate on the strategies employed for point sampling. 

\subsection{Different point trackers}

Our research builds upon recent advancements in point tracking methods, which excel in tracking points across frames within a video sequence. While our method is compatible with various point trackers, we primarily utilize CotrackerV2~\cite{cotracker} as our main point tracker. This choice is driven by its efficiency and capability to effectively track similar points together.

However, to demonstrate the flexibility of our approach with respect to the choice of point tracker, we present results using two additional recent trackers, TAPIR~\cite{doersch2023tapir} and PIPS++~\cite{zheng2023pointodyssey}, in Table \ref{table:tracker}. These trackers were selected because they both provide results for long-term tracking scenarios. We report our findings within the context of the 5-way k-shot setting across different datasets and k-shot settings.

Notably, our analysis reveals a consistent performance across the various trackers and settings. This substantiates our claim that our method is not tied to any specific point tracker and can effectively accommodate different trackers interchangeably.

\subsection{Point Sampling}
To effectively track points on objects appearing in subsequent frames, we employ a strategy of initializing points uniformly across multiple frames. This approach guarantees the tracking of points on objects that emerge after the initial frame. However, there are two notable considerations: 1) Redundancy may occur where a new point is sampled very close to an existing point and follows the same trajectory, thus not adding any new information. 2) The number of points may exceed the predetermined limit.

In the case of redundant points, when a point initialized at a later time is within a $\delta$ distance from an older point, and both points follow a similar trajectory—meaning all subsequent points also fall within the $\delta$ distance from each other—the older point is retained, while the new point is discarded.

After removing redundant points, the total number of points may still surpass the predefined limit, set to 256 in our experiments. To address this, we implemented various sampling strategies to reach this limit. Initially, we employed a straightforward random sampling method. Subsequently, we utilized two additional strategies based on the characteristics of the points and their trajectories: 1) sampling based on the length of the trajectory (Trajectory length), and 2) sampling based on the orientation of displacement within the trajectory (HoD).

\subsubsection{Trajectory length} 
For each point, we computed the length of its trajectory by calculating the Euclidean distance between consecutive points along the trajectory and summing these distances. Our hypothesis suggests that points with similar trajectory lengths might pertain to the same object, implying that some of these points could be redundant. To address this, we proceeded to calculate the trajectory lengths for all points and generated a histogram of these lengths. Subsequently, we employed stratified sampling, where roughly equal samples were selected from each histogram bin. This method aimed to discard points with similar trajectory lengths, optimizing the point selection process.

\begin{table*}[t!]
\begin{minipage}[b]{0.25\linewidth}
\addtolength{\tabcolsep}{10pt} 
\caption{Count of trainable parameters of previous methods.}
\label{table:params}
\resizebox{\linewidth}{!}{
\begin{tabular}{@{}lc@{}}

\toprule

\textbf{Methods} & \textbf{Params}\\
\midrule
OTAM~\cite{otam} & 23.5M\\
TRX~\cite{trx} & 47.1M\\
HYRSM~\cite{hyrsm} & 65.6M\\
MoLo~\cite{molo} & 89.6M\\
\textbf{Ours} & \textbf{10.8M}\\

\bottomrule
\end{tabular}}
\end{minipage}
\hfill
\begin{minipage}[b]{0.7\linewidth}
\centering

\addtolength{\tabcolsep}{1pt} 
\caption{Comparison of different point trackers for 1-shot, 3-shot, and 5-shot settings.}
\label{table:tracker}
\resizebox{\linewidth}{!}{
\begin{tabular}{@{}lccc|ccc|ccc@{}}
\toprule
&  \multicolumn{3}{c}{\textbf{Kinetics}} & \multicolumn{3}{c}{\textbf{SSv2-Full}}  & \multicolumn{3}{c}{\textbf{SSv2-Small}}  \\ \cmidrule(l){2-4} \cmidrule(l){5-7} \cmidrule(l){8-10} 
\textbf{Point tracker} &  \multicolumn{1}{l}{1-shot} & \multicolumn{1}{l}{3-shot} & \multicolumn{1}{l}{5-shot} & \multicolumn{1}{l}{1-shot} & \multicolumn{1}{l}{3-shot} & \multicolumn{1}{l}{5-shot} & \multicolumn{1}{l}{1-shot} & \multicolumn{1}{l}{3-shot} & \multicolumn{1}{l}{5-shot} \\
\midrule
TAPIR~\cite{doersch2023tapir} & 80.8 & 88.3 & 90.3 & 56.9 & 69.7 & 73.9 & 47.1 & 58.8 & 64.2\\
PIPS++~\cite{zheng2023pointodyssey} & 82.5 & 89.2 & 90.4 & 57.3 & 70.2 & 74.1 & 47.5 & 59.3 & 64.3\\
CotrackerV2~\cite{cotracker} & 81.9 & 89.9 & 91.1 & 57.7 & 70.0 & 74.6 & 47.9 & 60.0 & 64.4\\

\bottomrule
\end{tabular}}

\end{minipage}

\end{table*}

\subsubsection{HOD} Histogram of Oriented Displacements (HOD)~\cite{hod} is an extension of the Histogram of Oriented Gradients (HOG)~\cite{hog} method, designed for constructing descriptors for 2D trajectories. Similar to HOG, HOD considers the angle and magnitude of the displacement vector between two consecutive points, forming a descriptor where the magnitude serves as the value for the orientation bin.

In our approach, we create HOD descriptors for each trajectory using 8 orientation bins. After obtaining the HOD descriptors for all trajectories, we apply an agglomerative clustering~\cite{finch} technique to cluster these descriptors. Subsequently, we select an equal number of points from each cluster. The rationale behind this step is that points sharing similar HOD descriptors likely convey comparable information. Thus, by sampling from these clusters, we aim to discard redundant points while preserving essential trajectory information.

In Table \ref{table:sampling}, we present the 5-way k-shot performance results for three sampling strategies: trajectory length, HOD, and random sampling. Our observations across the three datasets indicate that these strategies yield comparable performance levels. Consequently, this work primarily focuses on reporting results for the random sampling strategy due to its simplicity. 

\begin{table}[t!]
\addtolength{\tabcolsep}{2pt} 
\caption{Comparison of different sampling techniques for 1-shot, 3-shot, and 5-shot settings.}
\label{table:sampling}
\resizebox{\linewidth}{!}{
\begin{tabular}{@{}lccc|ccc|ccc@{}}
\toprule
&  \multicolumn{3}{c}{\textbf{Kinetics}} & \multicolumn{3}{c}{\textbf{SSv2-Full}}  & \multicolumn{3}{c}{\textbf{SSv2-Small}}  \\ \cmidrule(l){2-4} \cmidrule(l){5-7} \cmidrule(l){8-10} 
\textbf{Sampling} &  \multicolumn{1}{l}{1-shot} & \multicolumn{1}{l}{3-shot} & \multicolumn{1}{l}{5-shot} & \multicolumn{1}{l}{1-shot} & \multicolumn{1}{l}{3-shot} & \multicolumn{1}{l}{5-shot} & \multicolumn{1}{l}{1-shot} & \multicolumn{1}{l}{3-shot} & \multicolumn{1}{l}{5-shot} \\
\midrule
Trajectory length & 82.1 & 90.2 & 91.0 & 57.4 & 69.9 & 74.3 & 47.4 & 59.9 & 64.1\\
HOD & 81.8 & 89.8 & 90.7 & 57.1 & 70.3 & 74.7 & 48.2 & 60.3 & 63.9\\
Random & 81.9 & 89.9 & 91.1 & 57.7 & 70.0 & 74.6 & 47.9 & 60.0 & 64.4\\

\bottomrule
\end{tabular}}
\end{table}







%

\end{document}